\documentclass[a4paper]{article}

\pdfoutput=1

\usepackage{INTERSPEECH2021}
\usepackage{kotex}
\usepackage{url}
\usepackage{graphicx}
\usepackage{multirow}
\usepackage{makecell}
\setcellgapes{2pt}

\title{kosp2e: Korean Speech to English Translation Corpus}
\name{Won Ik Cho$^1$, Seok Min Kim$^1$, Hyunchang Cho$^2$, Nam Soo Kim$^1$}
\address{
  $^1$Department of Electrical and Computer Engineering and INMC, Seoul National University\\
  $^2$PAPAGO, NAVER}
\email{\{wicho,smkim\}@hi.snu.ac.kr, hyunchang.cho@navercorp.com, nkim@snu.ac.kr}

\begin{document}

\maketitle
\begin{abstract}
 
Most speech-to-text (S2T) translation studies use English speech as a source, which makes it difficult for non-English speakers to take advantage of the S2T technologies. For some languages, this problem was tackled through corpus construction, but the farther linguistically from English or the more under-resourced, this deficiency and underrepresentedness becomes more significant. In this paper, we introduce \textit{kosp2e} (read as `kospi'), a corpus that allows Korean speech to be translated into English text in an end-to-end manner. We adopt open license speech recognition corpus, translation corpus, and spoken language corpora to make our dataset freely available to the public, and check the performance through the pipeline and training-based approaches. Using pipeline and various end-to-end schemes, we obtain the highest BLEU of 21.3 and 18.0 for each based on the English hypothesis, validating the feasibility of our data. We plan to supplement annotations for other target languages through community contributions in the future.

\end{abstract}

\section{Introduction}

Speech to text (S2T) translation is achieving the text of the target language from the speech of the source language. Speech translation is actively used in international conferences, video subtitling, and real-time translation for tourists. Traditionally, automatic speech recognition (ASR) and machine translation (MT) were adopted as a cascading method. Recently, end-to-end approaches have been widely utilized owing to the success of data-driven methodologies \cite{berard2016listen,berard2018end}.

However, studies on end-to-end speech translation using languages other than English as sources have not been driven actively. For instance, the survey in MuST-C \cite{di-gangi-etal-2019-must} suggests that among the 13 speech translation corpora, 12 handle English as source or target, and about half of the total temporal volume (547 hours over 1080 hours), including the largest ones \cite{niehues2018iwslt,kocabiyikoglu2018augmenting} only deals with English as a source speech. 

On one side, the reason English is exploited as a source language is probably that it fulfills large industrial needs \cite{nickerson2005english}. It is well-known as a world-widely used language that facilitates communication. Besides, recent S2T translation corpora \cite{kocabiyikoglu2018augmenting,di-gangi-etal-2019-must} usually leverages the ASR databases which were mainly proposed for English speech recognition, such as LibriSpeech \cite{panayotov2015librispeech}, which is plausible considering that recording speech in individual languages is extremely costly. 

However, speech translation shows a difference in application depending on what the source language is. In international conferences where speakers usually utter English, speech translation modules trained with an English source will obviously be effective. But what about YouTube video subtitling or tourists' real-time translation? Many videos contain utterances in the casts' own language. Also, it is challenging for non-English citizens to ask fluent English when they tour foreign countries. In other words, there are more points to consider on where non-English source speech can be utilized.

In this respect, we deemed that speech translation with a non-English speech source also needs attention. Among them, Korean is one of the languages that have been less investigated in S2T translation due to the deficiency of open spoken language corpora \cite{cho2020open} and its syntax and writing system being distinct from largely studied Indo-European (IE) languages. We construct a ko-en S2T translation dataset, \textit{kosp2e} (read as `kospi'), for the Korean spoken language, and release it publicly. This parallel corpus is the first to be constructed for S2T translation with Korean speech, and is constructed via manual translation of scripts, using or recording Korean speech databases.

Our contribution to the community is as follows:

\begin{itemize}
    \item We first build the Korean to English speech translation corpus, adopting the open-licensed speech recognition, machine translation, and spoken language corpora.
    \item We check that the feasibility of using ASR pretrained model in an end-to-end manner is promising in ko-en speech translation as well.
\end{itemize}

\section{Related Work}

Automatic speech recognition (ASR) denotes transcribing speech into text with the letters adopted in the language's writing system. Therefore, speech corpus used for training and evaluation of ASR has audio files and transcripts (sometimes with timestamps) accordingly. In the case of Korean speech recognition, Zeroth\footnote{\url{https://github.com/goodatlas/zeroth}}, KSS \cite{park2018kss}, and ClovaCall \cite{ha2020clovacall} are used as benchmark resources in the literature. 

Machine translation (MT) is mapping a text of the source language into the text of the target language. Due to this characteristic, MT corpora are often referred to as a parallel corpus. Storage and inspection of text corpora are easier than those of speech corpora, but bilingual processing is inevitable in making up MT corpora. This often makes construction costly, and there are not sufficient parallel corpora used as open resources in Korean \cite{cho2020open}. Although resources such as Korean Parallel Corpora \cite{park-etal-2016-korean} and Open Subtitles\footnote{\url{https://www.opensubtitles.org/ko}} exist, it is status quo that manually translated MT resources lack at this point compared to IE languages \cite{cho2020towards}. 

Speech translation is not simply a combination of speech recognition and machine translation database; it refers to a case in which tuples of source speech - source text - target text exist in parallel. The representative one is MuST-C \cite{di-gangi-etal-2019-must}, which contains the translation to eight languages with English TED talks as a source. It contains eight IE target language texts, namely German, Spanish, French, Italian, Dutch, Portuguese, Romanian, and Russian. There is recently distributed S2T dataset incorporating Korean regarding Duolingo challenge\footnote{\url{https://sharedtask.duolingo.com/}} \cite{staple20}, presented at the past IWSLT workshop. However, the collecting process was reading the prompts for fluency check rather than speaking natural language sentences, that we deemed difficult to utilize it for Korean train/test as our aim. Also, there is currently no domestic S2T translation resource available, up to our knowledge, whether it allows public access or not. 

\section{Corpus Construction}

We aim at the open-licensed distribution of our data contribution, either in commercial or non-commercial perspective. This allows the potential remix of the dataset, which can be done by not only the authors but also by community contribution. In this regard, we survey the open Korean spoken language corpora currently available with CC-BY license and list up the properties thereof, along with how we set up the principles in treating them in the recording and translation phase.

\subsection{Corpora}

\subsubsection{KSS}

KSS \cite{park2018kss} is a single speaker speech corpus used for Korean speech recognition and has a license of CC-BY-NC-SA 4.0\footnote{\url{https://www.kaggle.com/bryanpark/korean-single-speaker-speech-dataset}}. It consists of about 13K scripts, which a single voice actress utters and also includes the English translation. The scripts are sentences from the textbook for language education, and contain descriptions of daily life.

We re-recorded this corpus with verified crowd-workers so that diverse voices can be provided. In this process, no specific instruction was given. When it was not clear how to read, the participants were recommended to refer to the given translation or to record after listening to the source speech. A human inspector accompanied every output of the recording.

\subsubsection{Zeroth}

Zeroth is a Korean speech recognition dataset, and is released under a license of CC-BY 4.0\footnote{\url{https://github.com/goodatlas/zeroth}}. It consists of a total of about 22K scripts, and 3K unique sentences extracted from the news domain are used as scripts.

Zeroth was adopted for translation. Since it mainly consists of news scripts, it contains information on politics, social events, and public figures. We had the following guideline for the translation:

\begin{itemize}
    \item Some reporter names are inserted at the very first of the scripts. Translators may take it into account by, for example, placing a comma after the name coming at the front of the sentence.
    \item Public figures’ names are fixed to the format of ‘Family name - First name’ such as ‘Cho Won-ik’.
    \item Entities such as organizations, locations, or titles, are translated into English if adequate cognate exists, and are romanized if not.
\end{itemize}

\begin{table*}[]
\centering
\caption{\textit{kosp2e} subcorpus specification by domain.}
\resizebox{0.9\textwidth}{!}{%
\begin{tabular}{|c|c|c|c|c|c|c|}
\hline
\textbf{Dataset}  & \textbf{License} & \textbf{Domain}                                                                & \textbf{Characteristics}                                                                                                  & \textbf{\begin{tabular}[c]{@{}c@{}}Volume\\ (Train / Dev / Test)\end{tabular}}                                          & \textbf{\begin{tabular}[c]{@{}c@{}}Tokens\\ (ko / en)\end{tabular}} & \textbf{\begin{tabular}[c]{@{}c@{}}Speakers\\ (Total)\end{tabular}} \\ \hline
\textbf{Zeroth}   & CC-BY 4.0        & News / newspaper                                                               & \begin{tabular}[c]{@{}c@{}}DB originally for\\ speech recognition\end{tabular}                                            & \begin{tabular}[c]{@{}c@{}}22,247 utterances\\ (3,004 unique scripts)\\ (21,589 / 197 / 461)\end{tabular}               & 72K / 120K                                                          & 115                                                                 \\ \hline
\textbf{KSS}      & CC-BY-NC-SA 4.0  & \begin{tabular}[c]{@{}c@{}}Textbook\\ (colloquial\\ descriptions)\end{tabular} & \begin{tabular}[c]{@{}c@{}}Originally recorded\\ by a single speaker\\ (multi-speaker\\ recording augmented)\end{tabular} & \begin{tabular}[c]{@{}c@{}}25,708 utterances\\ = 12,854 * 2\\ (recording augmented)\\ (24,940 / 256 / 512)\end{tabular} & 128K / 190K                                                           & 17                                                                  \\ \hline
\textbf{StyleKQC} & CC-BY-SA 4.0     & \begin{tabular}[c]{@{}c@{}}AI agent\\ (commands)\end{tabular}                  & \begin{tabular}[c]{@{}c@{}}Speech act (4) \\ and topic (6)\\ labels are included\end{tabular}                             & \begin{tabular}[c]{@{}c@{}}30,000 utterances\\ (28,800 / 480 / 720)\end{tabular}                                        & 237K / 391K                                                         & 60                                                                  \\ \hline
\textbf{Covid-ED} & CC-BY-NC-SA 4.0  & \begin{tabular}[c]{@{}c@{}}Diary\\ (monologue)\end{tabular}                    & \begin{tabular}[c]{@{}c@{}}Sentences are in\\ document level;\\ emotion tags included\end{tabular}                        & \begin{tabular}[c]{@{}c@{}}32,284 utterances\\ (31,324 / 333 / 627)\end{tabular}                                        & 358K / 571K                                                         & 71                                                                  \\ \hline
\end{tabular}%
}

\label{tab:my-table}
\end{table*}

\subsubsection{StyleKQC}

StyleKQC \cite{cho2021stylekqc} is a script of 30K Korean directives that can be used in the AI agent domain, publicly available as CC-BY-SA 4.0\footnote{\url{https://github.com/cynthia/stylekqc}}. It contains six topics, namely messenger, calendar, weather and news, smart home, shopping, and entertainment, and four speech acts of alternative questions, \textit{wh-}questions, requirements, and prohibition. In addition, every utterance is tagged with either it has a formal or informal style of speaking. Ten sentences with the same intent are provided in a group. The original corpus contains text without punctuation, but in the translation and recording process, the speech act is also provided so that one can disambiguate the semantics and reflect it in translation or recording. Specifically, the following was suggested.

\medskip

\phantom{( )}\textit{Recording}
 \begin{itemize}
    \item Though the original script does not contain punctuation, the speech act types should be taken into account while recording.
    \item The tone of the recording should be differentiated between formal and informal utterances; formal utterances carefully as when facing elderly or colleagues, and informal utterances as when heading friends. 
    \item Albeit some sentences may seem awkward due to scrambling or fillers, read it as naturally as possible by placing a delay or pronouncing it chunk by chunk.
\end{itemize}

\medskip

\phantom{( )}\textit{Translation}
\begin{itemize}
    \item Although the sentences of the same intent may have the same connotation, the difference in style of each utterance should be reflected in the translation (e.g., "\textit{How many people are there in the US currently who are diagnosed as Covid 19?}" vs. "\textit{What is the current number of Covid 19 patients in the US?}" as a difference in the sentence structure, or "\textit{You know the current number of Covid 19 people in States?}" as a difference in the formality)
    \item Since the original corpus contains free-style spoken language that includes scrambling, filler, or sometimes fragments, the translators are asked to reflect that into the English translation
    \item Translation on names and entities follow the principles proposed for Zeroth.
\end{itemize}

\subsubsection{Covid-ED}

Covid-ED (COVID-19 Emotion Diary with Empathy and Theory-of-Mind Ground Truths Dataset) is a collection of crowdsourced diaries written in pandemic situations. It is labeled with emotions, empathy, personality, and levels of theory-of-mind. The dataset is publicly available as CC-BY-NC-SA 4.0\footnote{\url{https://github.com/humanfactorspsych/covid19-tom-empathy-diary}} \cite{lee_jung_lee_park_hahn_2021}. Upon writing diaries, workers were told  to either exclude or anonymize personally identifiable information (e.g., address, residential number, bank accounts, etc.) Each worker wrote five diaries for five consecutive days. Ground truth emotion labels per document were provided by the writers so that such information can be reflected in translation and recording. We considered the following in the process of recording and translation.

\medskip

\phantom{( )}\textit{Recording}
\begin{itemize}
\item One participant should record all the diaries of one writer, and in this process, the gender and age of the diary writer and those of the one who records should be aligned as much as possible.
\item The recording should be done considering (up to) two emotions that are tagged per diary. However, in this case, emotions do not need to be reflected in every sentence, instead, to relevant sentences. 
\item The diary is written in monologue format, but is basically web text, that there are various non-Hangul expressions in the script. Therefore, there may be variance in reading English words, numbers, and special symbols. The details of handling those are separately provided\footnote{\url{https://github.com/warnikchow/kosp2e/wiki/English,-Numbers,-and-Symbols-in-Covid-ED}}.
\end{itemize}

\medskip

\phantom{( )}\textit{Translation}
\begin{itemize}
    \item Sentences appear subsequently due to the script being a diary; thus, the subject omission may occur accordingly. In translating these parts, the translator should consider as much as possible the circumstances that the sentences are treated independently.
    \item English words should be translated as they are.
The numbers should be translated according to whether it is a cardinal/ordinal number. 
    \item Leetspeaks such as ㅋㅋ(laughter) and ㅠㅠ(sadness) should be translated to online expressions such as `lol’ and `T.T’.
\end{itemize}

\subsection{Recording and Translation}

\subsubsection{Recording}

The recording was conducted for KSS, StyleKQC, and Covid-ED.
KSS includes speech files, but since it was recorded by a single speaker, additional recordings were performed as mentioned above. Our final corpus includes both original and recorded versions. StyleKQC and Covid-ED were recorded in the same way. 
The recording was performed by selected workers in the crowd-sourcing group \textit{Green Web}\footnote{\url{https://www.gwebscorp.com/}} and \textit{Ever Young}\footnote{\url{http://everyoungkorea.com/}}, with some considerations regarding device and environment\footnote{\url{https://github.com/warnikchow/kosp2e/wiki/Considerations-in-Recording}}.


\subsubsection{Translation}

The translation was conducted for Zeroth, StyleKQC, and Covid-ED. All the texts were translated by a translation expert and checked by an inspector, managed by Lexcode\footnote{\url{https://lexcode.co.kr/}}. Except for KSS, the same post-processing rules were applied to scripts that have undergone expert translation (normalizing double spaces, removing non-Hangul and special tokens, etc.). In addition, we refined the translation of KSS referring to this standard.

\subsection{Statistics}

The specification of our corpus is as follows.

\begin{itemize}
    \item Utterances: 110,239 / Hours: 198H
    \item Source language (tokens): Korean (795K)
    \item Target language (tokens): English (1,272K)
    \item Speakers: 263 (in total)
    \item Domains: News, Textbook, AI agent command, Diary
\end{itemize}

Domain-wise descriptions of the subcorpora are in Table 1. For further usage, we attach the license policy of the original corpora. Two are non-commercial, but all subcorpora are available for remix and redistribution at least for academic purposes.  

\section{Experiment}

We split the proposed corpus into train, dev, and test set. In this process, for each type of subcorpus, speaker or sentence types were distributed with balance. In specific, for Zeroth, where the number of unique sentences is much smaller than that of the utterances, we reformulated the original train-test set so that there are only unseen sentences in the test set. For KSS, the original dataset was not included in the training set since the single speaker voices being dominant can badly affect the acoustic diversity of the corpus. For StyleKQC, where the original dataset has no particular test set for considering topic, act, and style at the same time, we split the whole dataset so that such attributes do not cause a bias. The split on Covid-ED was done in the way that there is no overlap of diary writers between each set.

In total, we have 1,266 and 2,320 utterances each for the dev and test set. The train set contains the rest, namely 106,653 utterances. We performed four ko-en speech translation experiments using the sets, with the evaluation metric of corpus BLEU \cite{papineni-etal-2002-bleu} of Sacrebleu \cite{post-2018-call} python library. Submodule performances were separately measured, especially using word error rate (WER) for Korean ASR. 

\begin{table}[]
\centering
\makegapedcells
\caption{Pipeline and end-to-end implementation using the constructed corpus. Note that the submodules of ASR-MT are evaluated with our test set, and those of ASR pretraining and warm-up are evaluated with randomly split validation set (not official).}
\resizebox{0.85\columnwidth}{!}{%
\begin{tabular}{|l|c|c|c|}
\hline
\multicolumn{1}{|c|}{\multirow{2}{*}{\textbf{Model}}} & \multirow{2}{*}{\textbf{BLEU}} & \multicolumn{2}{c|}{\textbf{Submodules}}                                                                                       \\ \cline{3-4} 
\multicolumn{1}{|c|}{}                                &                                & \textbf{\begin{tabular}[c]{@{}c@{}}WER\\ (ASR)\end{tabular}} & \textbf{\begin{tabular}[c]{@{}c@{}}BLEU\\ (MT/ST)\end{tabular}} \\ \hline
\textbf{ASR-MT (Pororo)}                              & 16.6                           & 34.0                                                         & 18.5 (MT)                                                       \\ \hline
\textbf{ASR-MT (PAPAGO)}                              & \textbf{21.3}                  & 34.0                                                         & 25.0 (MT)                                                       \\ \hline
\textbf{Transformer (Vanilla)}                        & 2.6                            & -                                                            & -                                                               \\ \hline
\textbf{ASR pretraining}                              & 5.9                            & 24.0*                                                        & -                                                               \\ \hline
\textbf{Transformer + Warm-up}                        & 11.6                           & -                                                            & 35.7 (ST)*                                                      \\ \hline
\textbf{\phantom{Transformer} + Fine-tuning}                                & \textbf{18.0}                  & -                                                            & -                                                               \\ \hline
\end{tabular}%
}
\label{tab:my-table}
\end{table}

\subsection{ASR-MT Pipeline}

In the ASR-MT pipeline, publicly available ASR and MT modules were used. We adopted Speech Recognition \cite{speechrecognition}\footnote{\url{https://github.com/Uberi/speech_recognition}} toolkit for ASR, using Korean as option, and MT was performed with Pororo\footnote{\url{https://github.com/kakaobrain/pororo}} \cite{pororo} NLP toolkit and PAPAGO translation API\footnote{\url{https://papago.naver.com/}}. The performance of both modules was checked with the utterances of the test set (Table 2).

\subsection{End-to-end Manner}

For end-to-end implementation, fairseq-S2T \cite{wang2020fairseqs2t} based on fairseq \cite{ott2019fairseq} was used. Three approaches were implemented; first vanilla transformer \cite{vaswani2017attention}, second using ASR pretrained model, and the last augmenting pseudo-gold translations for model warm-up. For the vanilla model, we stacked 12 transformers for the encoder and 6 for the decoder, both with 8 attention heads. For the second approach, we adopted a large-scale speech corpus (of 1,000H Korean utterances) publicly available at AI HUB\footnote{The dataset is downloadable from \url{https://aihub.or.kr/aidata/105}, but the detailed usage should refer to \url{https://github.com/sooftware/KoSpeech} for non-Korean citizens.} for ASR pretraining. The ASR module was trained based on fairseq script using source language text, and the soft label was concatenated with the decoder transformer trained upon it. The dimension of embeddings and dropouts were fixed to 256 and 0.1 each. For the last, which is inspired by \cite{pino2020self}, we machine-translated Korean scripts of AI HUB data with PAPAGO API. The transformer is first warmed up 8 epochs by large-scale pseudo-gold speech translation samples, and is further fine-tuned 30 epochs with our data. 

\subsection{Results}

The results of the pipeline and end-to-end baseline models are exhibited in Table 2. In the pipeline, though the translation scheme of the dataset used in the adopted module may differ from our translation scheme, we deemed that the results show the consistency of our corpus with conventional MT benchmarks used for Korean speech recognition and ko-en MT. In an end-to-end manner, we have obtained a mediocre performance for the vanilla transformer but a much-improved result for the ones using ASR pretrained model and pseudo-gold translations. The results suggest that it is not feasible with vanilla end-to-end models to yield transformation between Korean speech and English text at this point (which have different modality and distinct syntax at the same time), and it necessitates at least soft symbol-level representation or sufficient audio samples to get a satisfying ST performance.

\subsection{Discussion}

We intended to verify via an experiment that our corpus qualifies as a spoken language translation corpus. That is, we tried to show whether ours can reach the standard performance of ST benchmarks, overcoming the limitation in the gap of syntax and writing system between Korean and English. It seems that the fully end-to-end approach needs improvement, but the ones that leverage the ASR pretrained model or pseudo-gold translations promise the utility of advanced strategies such as large-scale ASR or ST pretrained accompanied with distillation strategies \cite{liu2019end}. 

Our corpus has both strengths and weaknesses. First, we have incorporated the various style of utterances, from formal (news/newspaper) to colloquial (daily descriptions/smart home/diary). Also, we considered the vocabularies of diverse domains, which may contribute to S2T translation in real-world applications. Lastly, we constructed our corpus leveraging open datasets so that the distributed version can be updated and redistributed with community contributions. This will make the dataset more viable to further annotations, such as tagging emotion or dialog acts. 

On the other side, our corpus has some weaknesses in its level of perfection, due to its scripts or recordings being adopted from open Korean language resources, which may have less consistency in between. Also, the recording format is not unified to a single type, which requires slightly more preprocessing (audio reading and resampling) before the training phase. However, the errata or other flaws in the original content were checked and corrected in the inspection process of our construction, and diverse speech files produced in non-fixed environments instead cover the speech inputs of various quality that are probable in real-world applications. Therefore, we deem that the weaknesses do not hinder our contribution as the first fully open resource in from-Korean speech translation, and we expect the issues to be resolved with user reports and remixes.

\section{Conclusion}

Through this study, we constructed a to-English translation dataset using Korean, which is less studied in speech translation as a source speech. In this process, ASR/MT corpora and spoken language corpora were used, either by means of recording and translation, to make up the whole corpora of about 110K translated utterances. In addition, we evaluated the training results of our corpus via pipeline and end-to-end manner, obtained the best BLEU of 21.3 and 18.0 for each, and discussed where the improvement should be made. We plan to open this corpus publicly to contribute to the speech and translation community\footnote{\url{https://github.com/warnikchow/kosp2e}}, and will expand it into a corpus considering more target languages, with further community contributions.

\section{Acknowledgements}

This work was supported by PAPAGO, NAVER Corp. The authors appreciate Hyoung-Gyu Lee, ‪Eunjeong Lucy Park, Jihyung Moon, and Doosun Yoo for discussions and support.‬  Also, the authors thank Taeyoung Jo, Kyubyong Park, and Yoon Kyung Lee for sharing the resources.

\bibliographystyle{IEEEtran}

\bibliography{mybib}


\end{document}